\def\year{2022}\relax
\documentclass[letterpaper]{article} 
\usepackage{aaai22}  
\usepackage{times}  
\usepackage{helvet}  
\usepackage{courier}  
\usepackage[hyphens]{url}  
\usepackage{graphicx} 
\usepackage{natbib}  
\usepackage{caption} 
\DeclareCaptionStyle{ruled}{labelfont=normalfont,labelsep=colon,strut=off} 
\usepackage{algorithm}
\usepackage{algorithmic}

\usepackage{newfloat}
\usepackage{listings}
\floatstyle{ruled}
\newfloat{listing}{tb}{lst}{}
\floatname{listing}{Listing}
\usepackage{amsmath}
\usepackage{amssymb}
\usepackage{booktabs}
\usepackage{multirow}
\title{Dead Pixel Test Using Effective Receptive Field}
\author{
    Bum Jun Kim,\textsuperscript{\rm 1}
    Hyeyeon Choi,\textsuperscript{\rm 1}
    Hyeonah Jang,\textsuperscript{\rm 1}
    Dong Gu Lee,\textsuperscript{\rm 1}
    Wonseok Jeong,\textsuperscript{\rm 2}
    Sang Woo Kim\textsuperscript{\rm 1}
}
\affiliations{

    \textsuperscript{\rm 1} Department of Electrical Engineering, Pohang University of Science and Technology\\
    \textsuperscript{\rm 2} Graduate School of Artificial Intelligence, Pohang University of Science and Technology\\
    \{kmbmjn, hyeyeon, hajang, dgleee, wonseok.jeong, swkim\}@postech.edu

}

\usepackage{bibentry}

\begin{document}

\maketitle

\begin{abstract}
    Deep neural networks have been used in various fields, but their internal behavior is not well known. In this study, we discuss two counterintuitive behaviors of convolutional neural networks (CNNs). First, we evaluated the size of the receptive field. Previous studies have attempted to increase or control the size of the receptive field. However, we observed that the size of the receptive field does not describe the classification accuracy. The size of the receptive field would be inappropriate for representing superiority in performance because it reflects only depth or kernel size and does not reflect other factors such as width or cardinality. Second, using the effective receptive field, we examined the pixels contributing to the output. Intuitively, each pixel is expected to equally contribute to the final output. However, we found that there exist pixels in a partially dead state with little contribution to the output. We reveal that the reason for this lies in the architecture of CNN and discuss solutions to reduce the phenomenon. Interestingly, for general classification tasks, the existence of dead pixels improves the training of CNNs. However, in a task that captures small perturbation, dead pixels degrade the performance. Therefore, the existence of these dead pixels should be understood and considered in practical applications of CNN.
\end{abstract}

\section{Introduction}
\label{sec:Introduction}

Deep neural networks have demonstrated remarkable performance in various fields such as object detection, semantic segmentation, and image classification \cite{ren2015faster,redmon2016you,long2015fully,ronneberger2015u,he2015delving,tan2019efficientnet}. The performance of deep neural networks varies depending on the architecture. State-of-the-art results have been obtained by designing large neural networks with increased depth \cite{he2016deep}, width \cite{zagoruyko2016wide}, and cardinality \cite{xie2017aggregated}.

In architecture design, one of the factors often considered is the receptive field \cite{araujo2019computing}. For example, if two $3\times3$ convolutions are applied, a resulting feature covers a $5\times5$ area. As such, the pixel-level area covered by a specific feature is called the theoretical receptive field. Meanwhile, effective receptive field was proposed by \citet{luo2016understanding}. Contrary to the previous square-shaped theoretical receptive field, the effective receptive field illuminates the actually activated pixels through gradients, whose shape appears as a 2D Gaussian.

Many studies have preferred enlarging the receptive fields to obtain performance gain \cite{tsai2018learning,fu2018deep,singh2018analysis,kim2016accurate,johnson2016perceptual,shi2020pv,plotz2018neural}. Further, \citet{araujo2019computing} conjectured the relationship between the size of the receptive field and the classification accuracy. However, this study points out that these common practices should be reconsidered. For modern convolutional neural networks (CNNs), we measured the size of the receptive field. We observed that the large receptive field cannot guarantee the performance superiority of the neural network. For example, some neural networks exhibit high accuracy but have a smaller receptive field. This is because the size of the receptive field reflects only the depth or kernel size and does not reflect the width or cardinality.

In addition to examining the size of the effective receptive field, we investigate the shape of the effective receptive field. We further obtained the effective receptive field of the final output. Conventionally, for a CNN, every pixel, or at least an adjacent pixel, is expected to contribute almost equally to the final output. In other words, it would be strange if a pixel at a specific location is partially dead, and the dead pixel has little effect on the output for any data. Surprisingly, we found that the partially dead pixels exist. In modern CNNs such as ResNet, strong pixels and weak pixels exist for any data. Here, the contribution to the output is significantly different for strong pixels and weak pixels. We will show that this pixel sensitivity imbalance is significant even for adjacent pixels. This pixel sensitivity imbalance occurs when an operation with an odd-sized kernel is applied with stride 2. A solution to this problem is provided in Section \ref{sec:Shape_Test_on_the_Effective_Receptive_Field}.

Is pixel sensitivity imbalance a bug or a feature? We compared the performance of CNNs after reducing the pixel sensitivity imbalance. Interestingly, pixel sensitivity imbalance does not degrade but rather enhances the neural network's performance. In this respect, pixel sensitivity imbalance is a feature for general vision tasks. However, when pixel sensitivity imbalance exists, it is difficult to capture small perturbations in images. In contrast, when the pixel sensitivity imbalance is reduced, the neural networks easily capture small perturbations in images. In this regard, pixel sensitivity imbalance is a bug for some special tasks.

\section{Preliminaries: Receptive Field}
\label{sec:Background_Receptive_Field}

In the theoretical receptive field, the largest pixel-level area covered by the target feature is investigated by tracing backward operations in the CNN. For example, if three $3\times3$ convolutions are applied, one target feature has a $7\times7$ theoretical receptive field. If any of these operations have a stride greater than 1, the target feature will cover a larger area, resulting in a wider theoretical receptive field \cite{araujo2019computing}.

However, as the theoretical receptive field is the theoretical maximum area covered by a target feature, it is far from the practical behavior of the neural network. In the effective receptive field, gradients are used to examine the actual pixels that affect the target feature. Contrary to the theoretical receptive field that appears as a square, the effective receptive field appears as a 2D Gaussian.

Here we provide a detailed formulation of our trick to obtain the effective receptive field. Suppose an image $I_{xyz} \in \mathbb{R}^{224\times224\times3}$ is given. The image is passed through given CNN, resulting in a target feature map $A_{ijk} \in \mathbb{R}^{7\times7\times N_c}$. For effective receptive field, the goal is to represent the spatial relationship between pixel-level $(x, y)$ and feature-level $(i, j)$. Therefore, the channel of the image and feature map should be ignored and averaged. First, we define,
\begin{align}
	F=\frac{1}{N_c}\sum_k{A_{44k}},
\end{align}
which is the averaged feature over channel $k$ for the spatial center $(4, 4)$ in the target feature map. Then we compute the gradient w.r.t image, $\frac{\partial F}{\partial I_{xyz}}$. By averaging the gradient over the channel $z$, we obtain,
\begin{align}
	G_{xy}=\frac{1}{3}\sum_z{\frac{\partial F}{\partial I_{xyz}}},
\end{align}
which represents how pixel $(x, y)$ affects the central feature for the given image. However, the $G_{xy}$ from a single image is sparse and depends on the image. By averaging $G_{xy}$ over a sufficiently large number of data, the nature of the neural network can be obtained.

However, if some $G_{xy}$ has a negative value, it cancels out with a positive $G_{xy}$. As we want to obtain the accumulation of pixel contributions, we ignore negative importance \cite{selvaraju2017grad,chattopadhay2018grad}. Thus, we pass $G_{xy}$ through the $ReLU$ \cite{glorot2011deep}:
\begin{align}
	R_{xy} = \frac{1}{N}\sum_n{ReLU(G_{xy})}.
\end{align}

Now, $R_{xy}$ represents the general contribution property of pixel $(x, y)$ to the target feature, i.e., the effective receptive field. In summary, we need first to calculate $G_{xy}$ for each image, pass it through $ReLU$, and then average it over a sufficiently large dataset. In the modern deep learning environment using mini-batch, applying $ReLU$ to the gradient for each image can be difficult. We recommend using batch size 1 to correctly accumulate $ReLU(G_{xy})$ for each image. Accumulating $ReLU(G_{xy})$ over a sufficiently large amount of data yields a clean, high-quality effective receptive field $R_{xy}$ that well describes the internal behavior of a neural network.

\section{Size Test on the Effective Receptive Field}
\label{sec:Size_Test_on_the_Effective_Receptive_Field}

\begin{table*}[t!]
	\centering
	\begin{tabular}{l|l|l|l|l|l|l}
		\toprule
        \textbf{Model}    & \textbf{Acc@1} & \textbf{Acc@5} & \textbf{TRF} & $\hat\sigma_X$ & $\hat\sigma_Y$ & $R^2$ \\
		\midrule
		ResNet-18         & 69.758         & 89.078         & 435          & 76.534         & 73.662         & 0.908 \\
		ResNet-34         & 73.314         & 91.420         & 899          & 96.148         & 97.242         & 0.785 \\
		ResNet-50         & 76.130         & 92.862         & 427          & 64.820         & 62.208         & 0.945 \\
		ResNet-101        & 77.374         & 93.546         & 971          & 60.061         & 60.679         & 0.937 \\
		ResNeXt-50-32x4d  & 77.618         & 93.698         & 427          & 70.921         & 67.679         & 0.908 \\
		ResNet-152        & 78.312         & 94.046         & 1451         & 60.589         & 58.738         & 0.930 \\
		Wide-ResNet-50-2  & 78.468         & 94.086         & 427          & 58.415         & 60.676         & 0.954 \\
		Wide-ResNet-101-2 & 78.848         & 94.284         & 971          & 56.900         & 59.888         & 0.944 \\
		ResNeXt-101-32x8d & 79.312         & 94.526         & 971          & 77.664         & 77.047         & 0.889 \\
		\bottomrule
	\end{tabular}
\caption{For ResNet and its variants, we summarize top-k classification accuracy (\%), theoretical receptive field (TRF) size, effective receptive field size ($\hat\sigma_X$, $\hat\sigma_Y$), and $R^2$ from fitting. Contrary to previous studies, we observed that the classification accuracy was not proportional to the size of the theoretical receptive field. The size of the effective receptive field also did not show a tendency consistent with the classification accuracy.}
	\label{tab:rf_for_imagenet}
\end{table*}

\begin{table*}[t!]
	\centering
    \begin{tabular}{l|l|l|l|l|l|l}
		\toprule
        \textbf{Model}    & \textbf{Valid Acc} & \textbf{Test Acc} & \textbf{TRF} & $\hat\sigma_X$ & $\hat\sigma_Y$ & $R^2$ \\
		\midrule
		ResNet-18         & 93.00             & 91.83            & 435          & 68.439         & 71.996         & 0.953 \\
		ResNet-34         & 93.95             & 92.71            & 899          & 78.291         & 81.658         & 0.907 \\
		ResNet-101        & 94.38             & 92.71            & 971          & 57.916         & 60.300         & 0.933 \\
		Wide-ResNet-50-2  & 94.97             & 94.31            & 427          & 57.862         & 63.025         & 0.963 \\
		ResNet-50         & 94.82             & 94.38            & 427          & 56.528         & 59.245         & 0.962 \\
		ResNet-152        & 95.70             & 94.60            & 1451         & 58.907         & 60.345         & 0.916 \\
		ResNeXt-50-32x4d  & 95.04             & 94.97            & 427          & 61.416         & 63.959         & 0.951 \\
		Wide-ResNet-101-2 & 95.33             & 94.97            & 971          & 60.325         & 64.184         & 0.935 \\
		ResNeXt-101-32x8d & 96.50             & 95.40            & 971          & 64.720         & 67.788         & 0.912 \\
		\bottomrule
	\end{tabular}
    \caption{After fine-tuning each model on the Caltech-101 dataset, the same measurement was performed. Similarly, the classification accuracy (\%) did not show a tendency consistent with the size of the receptive field.}
	\label{tab:rf_for_caltech}
\end{table*}

Here, we investigate the size of the effective receptive field of modern CNNs. Target CNNs are ResNet and its variants \cite{he2016deep,zagoruyko2016wide,xie2017aggregated}, which are widely used in various vision tasks. We used \texttt{torchvision.models} \cite{paszke2019pytorch,paszke2017automatic} that were pre-trained on ImageNet \cite{russakovsky2015imagenet}. For each model, we summarized the top-1 accuracy and top-5 accuracy reported. We also computed the size of the theoretical receptive field for each model. Note that since ResNet differs in detailed architecture for each implementation, the sizes of the theoretical receptive field for \texttt{torchvision.models} are different from that of the \texttt{TensorFlow} models \cite{araujo2019computing,silberman2016tensorflow}.

For each pre-trained model, we obtained an effective receptive field using the test dataset from the CUB-200-2011 dataset \cite{wah2011caltech}. Here, we set the target feature map as the last feature map of $7\times7$ layer-4. For the effective receptive field using $14\times14$ layer-3 as the target feature map, refer to the supplementary material. The obtained effective receptive field was fitted with 2D Gaussian using the \texttt{Lmfit} library \cite{newville2016lmfit}. The resulting $\hat\sigma_X$ and $\hat\sigma_Y$ indicate how large the effective receptive field is. These results are summarized in Table \ref{tab:rf_for_imagenet}. Our major observations are summarized as follows.
\paragraph{Observation 1. The size of the theoretical receptive field does not describe the classification accuracy.}
We observed that the classification accuracy of CNNs is not proportional to the size of the theoretical receptive field. For example, Wide-ResNet-50-2 and ResNet-152 show similar classification accuracy, but the size of their theoretical receptive field is 427 and 1451 pixels, respectively. This is because the theoretical receptive field reflects only depth or kernel size and cannot reflect width or cardinality. On the other hand, ResNet-34 has a large theoretical receptive field of 899 pixels because it uses early convolution with stride 2 in residual blocks, unlike ResNet-50. As such, ResNet-34 has a wider theoretical receptive field than ResNet-50, but its classification accuracy is lower. These observations are inconsistent with the conjecture \cite{araujo2019computing} that the classification accuracy tends to be proportional to the size of the theoretical receptive field.

\paragraph{Observation 2. The size of the effective receptive field does not describe the classification accuracy.}
We observed that the classification accuracy of CNNs is also not proportional to the size of the effective receptive field. In other words, even from the viewpoint of the effective receptive field, the large receptive field does not guarantee superiority in performance. Meanwhile, when the depth increases within these ResNets, the size of the effective receptive field does not increase further and saturates to a certain size. These results are different from the study of \citet{luo2016understanding}, which reported that the size of the effective receptive field tends to be proportional to $\sqrt{depth}$.

The same experiment was performed once more. First, each pre-trained ResNet was fine-tuned on the Caltech-101 dataset \cite{fei2004learning}. We replaced the last fully connected layer to output for 101 classes. For training, stochastic gradient descent with momentum 0.9 \cite{sutskever2013importance}, learning rate 0.01, weight decay 0.0005, batch size 64, epochs 200, and cosine annealing schedule with 200 iterations \cite{loshchilov2016sgdr} was used. For data augmentation, random resized crop with size 256, random rotation with degree 15, color jitter, random horizontal flip, center crop with size 224, and mean-std normalization was applied. The train/val/test set was split at a ratio of 70:15:15. Within 200 epochs, the model with the best validation accuracy was obtained and evaluated.

For each fine-tuned model, an effective receptive field was obtained using the test dataset, and its size was investigated (Table \ref{tab:rf_for_caltech}). Similarly, the size of the theoretical receptive field and the effective receptive field do not agree with the trends in classification accuracy.  Therefore, we conclude that the size of the receptive field is not a representative indicator of classification accuracy, nor architectural superiority.

\section{Shape Test on the Effective Receptive Field}
\label{sec:Shape_Test_on_the_Effective_Receptive_Field}

In the previous section, we fitted each effective receptive field to a 2D Gaussian. Although $R^2$ showed near 0.9, those effective receptive fields did not perfectly match the 2D Gaussian. To understand this behavior, we visualize the obtained effective receptive field.

\begin{figure*}[t!]
	\centering
	\includegraphics[width=0.3\textwidth]{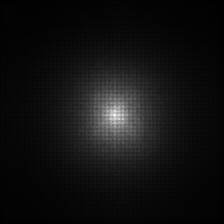}
	\includegraphics[width=0.3\textwidth]{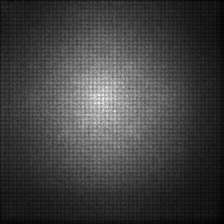}
	\includegraphics[width=0.3\textwidth]{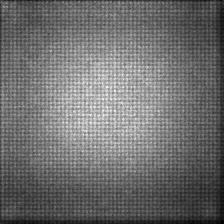}
	\includegraphics[width=0.3\textwidth]{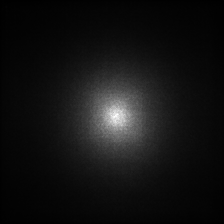}
	\includegraphics[width=0.3\textwidth]{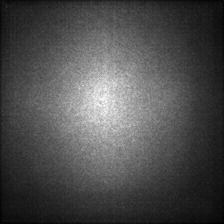}
	\includegraphics[width=0.3\textwidth]{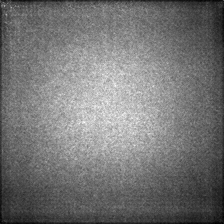}
	\caption{Effective receptive fields were obtained for ResNeXt-101-32x8d. (Left) Effective receptive field of the $7\times7$ feature map. (Middle) Effective receptive field of the $14\times14$ feature map. (Right) Effective receptive field of the output. The top row is the effective receptive fields for the existing ResNeXt-101-32x8d before kernel padding, showing the checkboard pattern. The bottom row is the effective receptive fields after kernel padding, showing no checkboard pattern. Best viewed electronically with zoom.}
	\label{fig:erf}
\end{figure*}

For the ResNeXt-101-32x8d, we plotted the effective receptive field (Figure \ref{fig:erf}). Although the effective receptive field appears as 2D Gaussian, a checkboard pattern exists inside. Therefore, the effective receptive field imperfectly matched the 2D Gaussian because of the internal checkboard pattern.

Additionally, we accumulated $\frac{\partial y}{\partial I}$ to obtain an effective receptive field of output $y$. This is what we call the dead pixel test. In general, the entire pixels are expected to contribute almost equally to output $y$. However, even for the effective receptive field of $y$, we discovered that the checkboard pattern exists.

The existence of this checkboard pattern implies that modern CNNs recognize images in a highly counterintuitive way. Some pixels are weak, partially dead, and hardly contribute to the output. Conversely, some pixels are strong and more sensitive to output. We call this phenomenon \textit{pixel sensitivity imbalance}. As the checkboard pattern appears locally, even in adjacent pixels, the pixel sensitivity differs significantly.

\begin{figure}[t!]
	\centering
	\includegraphics[width=0.85\columnwidth]{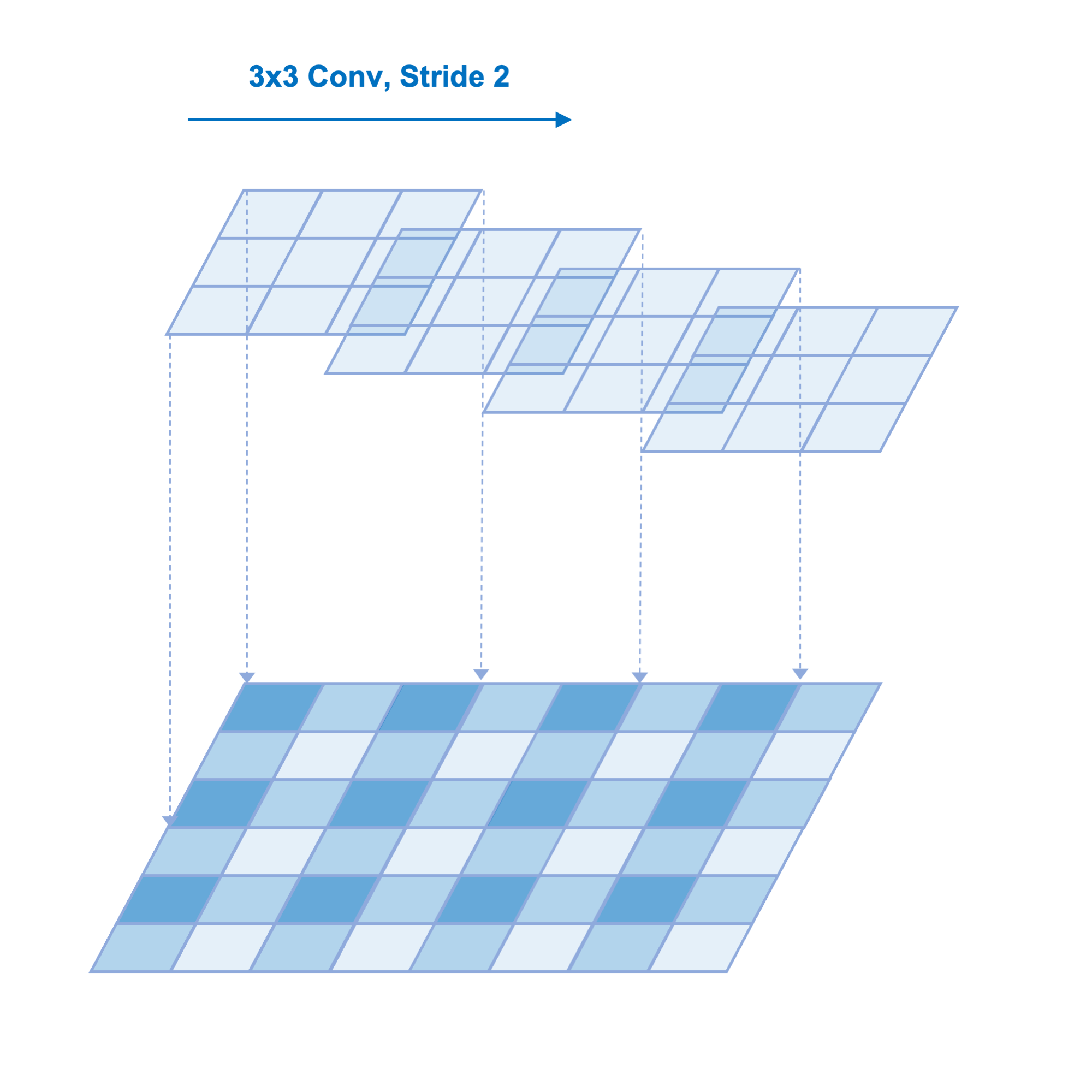}
	\caption{When an odd-sized kernel is applied with stride 2, a checkboard pattern appears.}
	\label{fig:checkboard}
\end{figure}

Why does the checkboard pattern appear? We found that it occurs when an odd-sized kernel is applied with stride 2 (Figure \ref{fig:checkboard}). For example, when a $3\times3$ convolution is applied with stride 2, overlapping regions appear. Pixels within the overlapping regions are referenced more in operation, while other pixels are not. As this phenomenon accumulates, some pixels become more influential while others do not. When viewed in 2D, a checkboard pattern appears. This phenomenon is highly similar to the checkboard pattern when using deconvolution in image generation tasks \cite{odena2016deconvolution}. Extending this, we emphasize that the checkboard pattern exists from the perspective of gradient even when using convolution.

Despite these potential problems, odd-sized kernels with stride 2 are widely used in modern CNNs \cite{huang2017densely,szegedy2015going,iandola2016squeezenet,krizhevsky2012imagenet,ma2018shufflenet,sandler2018mobilenetv2}. For example, in the early stage of ResNets, $7\times7$ Conv with stride 2 and $3\times3$ Pool with stride 2 are used. Further, in the downsampling operation in the residual block, $1\times1$ Conv is used with stride 2, which subsamples only the specific input and shuts off the flow in other locations.

Here, we would like to modify those problematic odd-sized kernels with stride 2. As ResNet and its variants have similar architectures, most can be modified with similar rules. Not all layers need to be modified. The operations to be modified are as follows: $7\times7$ Conv with stride 2 and $3\times3$ Pool with stride 2 in early stage, and $1\times1$ Conv with stride 2 and $3\times3$ Conv with stride 2 across all residual blocks. We replace those kernels with even sizes such as $8\times8$ or $4\times4$.

\begin{figure}[t!]
	\centering
	\includegraphics[width=0.9\columnwidth]{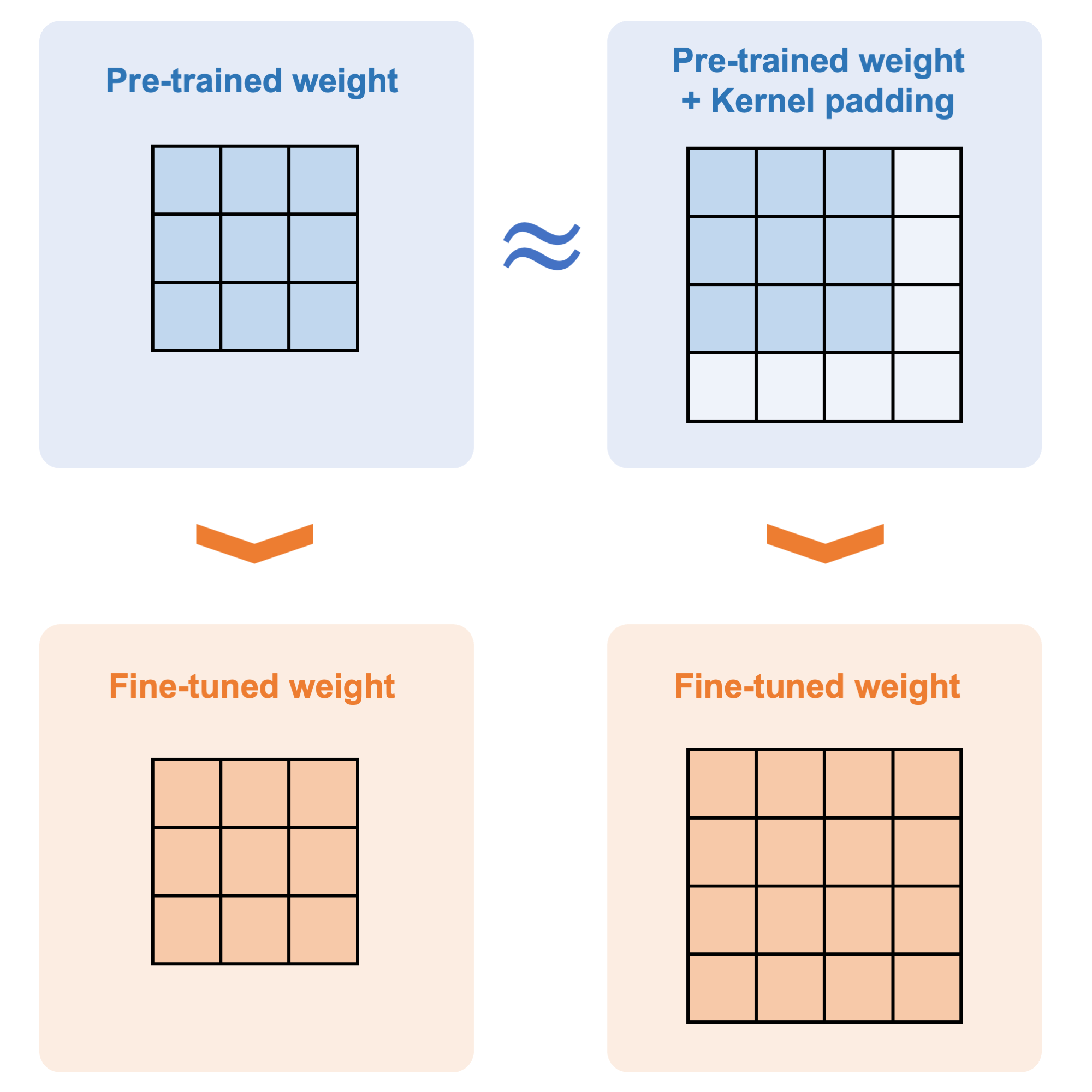}
	\caption{To construct an even-sized kernel while using pre-trained weights, we propose a kernel padding method. An even-sized kernel is constructed by applying zero-padding to the bottom and right sides of the kernel. After fine-tuning, the added weights along with the existing weights can be properly trained.}
	\label{fig:kp}
\end{figure}

However, when replacing with a new kernel, the existing pre-trained weights are discarded. To construct an even-sized kernel while boosting training through pre-trained weights, we propose \textit{kernel padding} method. For the target odd-sized pre-trained weight, zero-padding is applied to the bottom and right sides to obtain an even-sized kernel (Figure \ref{fig:kp}). As the kernel is zero-padded, the operation is equivalent to the previous one. Accordingly, pre-trained weights can be enjoyed. Moreover, as the new zero-padded weights are trainable, during fine-tuning, they can be merged into the existing weights.

We applied kernel padding to the ResNets pre-trained on the ImageNet and then fine-tuned them on the Caltech-101 dataset. The training details used in fine-tuning are the same as the experiments in Section \ref{sec:Size_Test_on_the_Effective_Receptive_Field}. Now the effective receptive field of our architecture has no checkboard pattern (Figure \ref{fig:erf}).

The degree of the pixel sensitivity imbalance can be measured through the smoothness of the effective receptive field $R_{xy}$ of output $y$. Here, we define two indices, first-order imbalance index $L_1$ and second-order imbalance index $L_2$:
\begin{align}
	L_1 = \frac{1}{224 \cdot 223 \cdot 2} \sum_{x,y}{(|\partial_x R_{xy}| + |\partial_y R_{xy}|)}, \\
	L_2 = \frac{1}{224 \cdot 222 \cdot 2} \sum_{x,y}{(|\partial_x^2 R_{xy}| + |\partial_y^2 R_{xy}|)}.
\end{align}

In other words, we pass the effective receptive field through the difference filters and compute spatial average to evaluate its local variation and curvature. The smaller these values are, the more locally smooth the effective receptive field is. Conversely, the larger the value, the greater the imbalance.

\begin{figure*}[t!]
	\centering
	\includegraphics[width=0.95\textwidth]{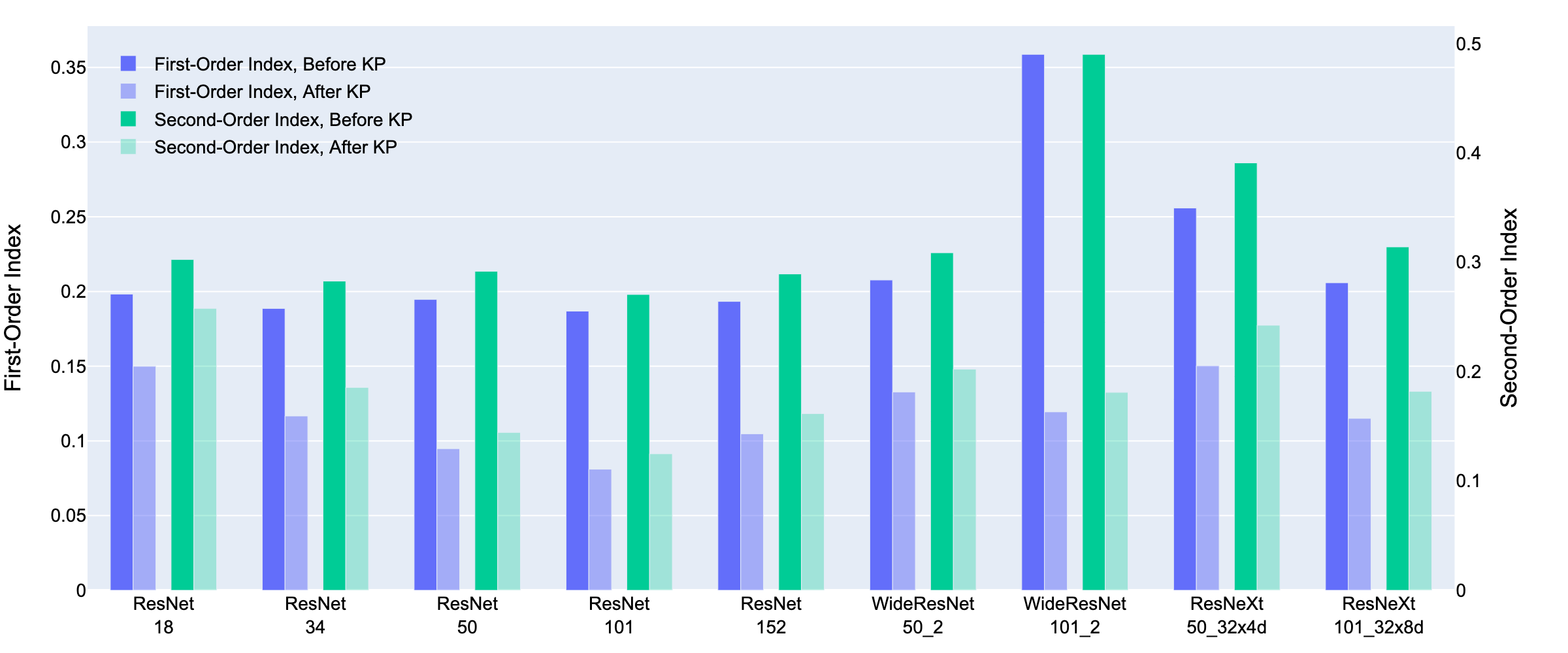}
	\caption{To quantitatively evaluate pixel sensitivity imbalance, we measured the two indices. In all the target architectures, pixel sensitivity imbalance is reduced after kernel padding.}
	\label{fig:index_kp}
\end{figure*}

Using these two indicators, we evaluated the degree of pixel sensitivity imbalance before and after applying kernel padding (Figure \ref{fig:index_kp}). Existing ResNets show large $L_1$ and $L_2$, which indicates that pixel sensitivity imbalance is significant even in adjacent pixels. After applying the kernel padding, the imbalance decreased across all ResNets.

\section{Discussion}
\label{sec:Discussion}

Is pixel sensitivity imbalance a bug or a feature? In other words, if the pixel sensitivity imbalance is reduced, can the superiority of the architecture be guaranteed? For both perspectives, we provide some conjectures.

\subsection{Pros: Pixel Sensitivity Imbalance is a Feature.}
\label{sec:Pros_Pixel_Sensitivity_Imbalance_is_a_Feature}

Even with pixel sensitivity imbalance, ResNets have been widely used in various vision tasks so far. Although some pixels are partially dead, they are not entirely dead. The difference between strong and weak pixels is a matter of contribution degree, and they are all involved in the output.

The phenomenon of blocking the flow of certain inputs is not so unfamiliar. For example, consider Dropout \cite{srivastava2014dropout} or DropBlock \cite{ghiasi2018dropblock}. They improve the performance of neural networks by dropping some neurons or inputs. For understanding the global context of an image, it is fine if some trivial input is missing. Furthermore, dropping some input induces the CNN to understand the image in a different way, introducing a regularization effect.

Further, pixel sensitivity imbalance can be interpreted as rescaling a given image according to strong and weak pixels. As the image is rescaled pixel-wise, when a translated image is given, it is recognized as a completely different image. Accordingly, pixel sensitivity imbalance increases image diversity, thereby boosting the effect of data augmentation.

Here, we examined how pixel sensitivity imbalances affect the performance in a general vision task. As kernel padding reduces pixel sensitivity imbalance, we compared the performance of ResNet and its variants before and after applying kernel padding. We performed fine-tuning on the Caltech-101 dataset, and the experimental details such as the training method and data augmentation are the same as in Section \ref{sec:Size_Test_on_the_Effective_Receptive_Field}.

\begin{table}[t!]
	\centering
	\begin{tabular}{l|r|r|r}
		\toprule
		\textbf{Model}    & \textbf{Before KP} & \textbf{After KP} & \textbf{Diff} \\
		\midrule
        ResNet-101        & \textbf{95.28}             & 94.07            &  1.22 $\downarrow$      \\
		Wide-ResNet-101-2 & \textbf{95.14}             & 94.09            &  1.05 $\downarrow$      \\
		ResNeXt-101-32x8d & \textbf{95.58}             & 94.75            &  0.83 $\downarrow$      \\
		\bottomrule
	\end{tabular}
    \caption{To investigate whether pixel sensitivity imbalance helps training or not, we compared the test accuracy (\%) before and after applying kernel padding. After kernel padding, the performance is rather decreased. Thus, for general vision tasks, pixel sensitivity imbalance is not a bug, and it is a feature.}
	\label{tab:general_task}
\end{table}

For each model, we measured the average test accuracy from three experiments (Table \ref{tab:general_task}). We observed that the performance is rather decreased after kernel padding. This means that pixel sensitivity imbalance is not a bug for a general image classification task but is a feature that improves performance. Therefore, reducing pixel sensitivity imbalance does not guarantee architectural superiority.

\subsection{Cons: Pixel Sensitivity Imbalance is a Bug.}
\label{sec:Cons_Pixel_Sensitivity_Imbalance_is_a_Bug}

Nevertheless, pixel sensitivity imbalance gives rise to several potential problems.

First, consider the saliency methods that visualize the inner behavior of a neural network. Many saliency methods have investigated important pixels based on gradients \cite{simonyan2013deep,springenberg2014striving,smilkov2017smoothgrad,sundararajan2017axiomatic,shrikumar2017learning}. However, saliency methods do not reflect pixel sensitivity imbalance. In other words, the gradient-based saliency map is affected by the checkboard pattern of the neural network. Thus, the gradient-based saliency method is only suitable for examining the pixels that contribute to the output of the neural network and is unsuitable for evaluating the intrinsic importance of a pixel.

As mentioned earlier, since pixel sensitivity imbalance introduces pixel-wise rescaling, the translated image is perceived as a completely different image. This increases the data augmentation effect but worsens the translation invariance of CNN \cite{zhang2019making,azulay2018deep,cohen2016group}. In a practical application, for example, if 1-pixel translated image produces a different result, the vision system would be considered unreliable and unstable.

Moreover, pixel sensitivity imbalance implies a positional difference for capturing a perturbation. Consider one-pixel attack \cite{su2019one}, which attempts an adversarial attack to invert the output by perturbing a certain pixel. Here we can additionally exploit the fact that strong pixels are generally more sensitive. If we construct an attack strategy that focuses more on strong pixels, we can attack the neural network more easily.

Here, we provide a mathematical formulation. Consider output $y$, the logit before softmax layer. For a CNN with $ReLU$-like activations, we can represent the output using piece-wise linear function \cite{srinivas2018knowledge,simonyan2013deep}:
\begin{align}
	y & = \frac{\partial y}{\partial I_{111}} I_{111} + ... + \frac{\partial y}{\partial I_{224,224,3}} I_{224,224,3} + C \\
	  & = \sum_{x,y,z}{\frac{\partial y}{\partial I_{xyz}} I_{xyz}} + C.
\end{align}

$\frac{\partial y}{\partial I_{xyz}}$ and $C$ are evaluated at specific image $I$. Here, we approximate them using
\begin{align}
	\frac{\partial y}{\partial I_{xyz}} & \approx R_{xy}, \\
	C                                   & \approx E(C),
\end{align}
which results in a fixed linear model, obtained from the mean over images. We define $\tilde{y} (I)$,
\begin{align}
	\tilde{y} (I) = \sum_{x,y}{ R_{xy} (I_{xy1} + I_{xy2} + I_{xy3})} + E(C),
\end{align}
which is the output from the fixed linear model.

Now, assume that we put a perturbation $\epsilon$ on $I_{XYZ}$. Then,
\begin{align}
	\Delta{\tilde{y}} & = \tilde{y} (I+\epsilon e_{XYZ}) - \tilde{y} (I) \\
                      & = \epsilon R_{XY}. \label{eq:pert}
\end{align}

If pixel sensitivity imbalance exists, $R_{XY}$ differs depending on the $(X, Y)$. Thus, even if the same amount of $\epsilon$ is applied, $\Delta{\tilde{y}}$ varies depending on where the perturbation is applied. For example, if we put perturbation to a strong pixel, the output can be significantly affected.

In addition, Eq. \ref{eq:pert} implies that when pixel sensitivity imbalance exists, it may be difficult to distinguish whether the change in output is due to the magnitude of the perturbation or the position of the perturbation. Then, if perturbations with different magnitudes are applied at random locations, can CNN distinguish the magnitudes of perturbations? Further, if the perturbation magnitude and the position are also randomly varied every time, and only the average of the perturbation magnitude has a difference, it will be quite a challenging problem. However, these problems are commonly encountered in practical vision tasks.

Here, we propose a \textit{micro-object classification task}. The templates are images from the Caltech-101 dataset. First, we select a random $8\times8$ region within the template. After changing the color of the selected area to RGB=(0, 0, 0), we label the image as class A. In the same way, for class B, select a random area, but replace it with RGB=(255, 0, 0). As such, we put a micro-object at a random location in the image to perform a binary classification task. In this task, not only the position of the perturbation but also the magnitude changes every time. Here, when pixel sensitivity imbalance exists, it may be difficult for CNN to capture the difference in perturbation magnitude between the two classes. In contrast, suppose the pixel sensitivity imbalance is reduced by kernel padding. In that case, as the influence of the random position decreases, the change in the magnitude of perturbation can be more easily captured.

\begin{figure}[t]
	\centering
	\includegraphics[width=0.9\columnwidth]{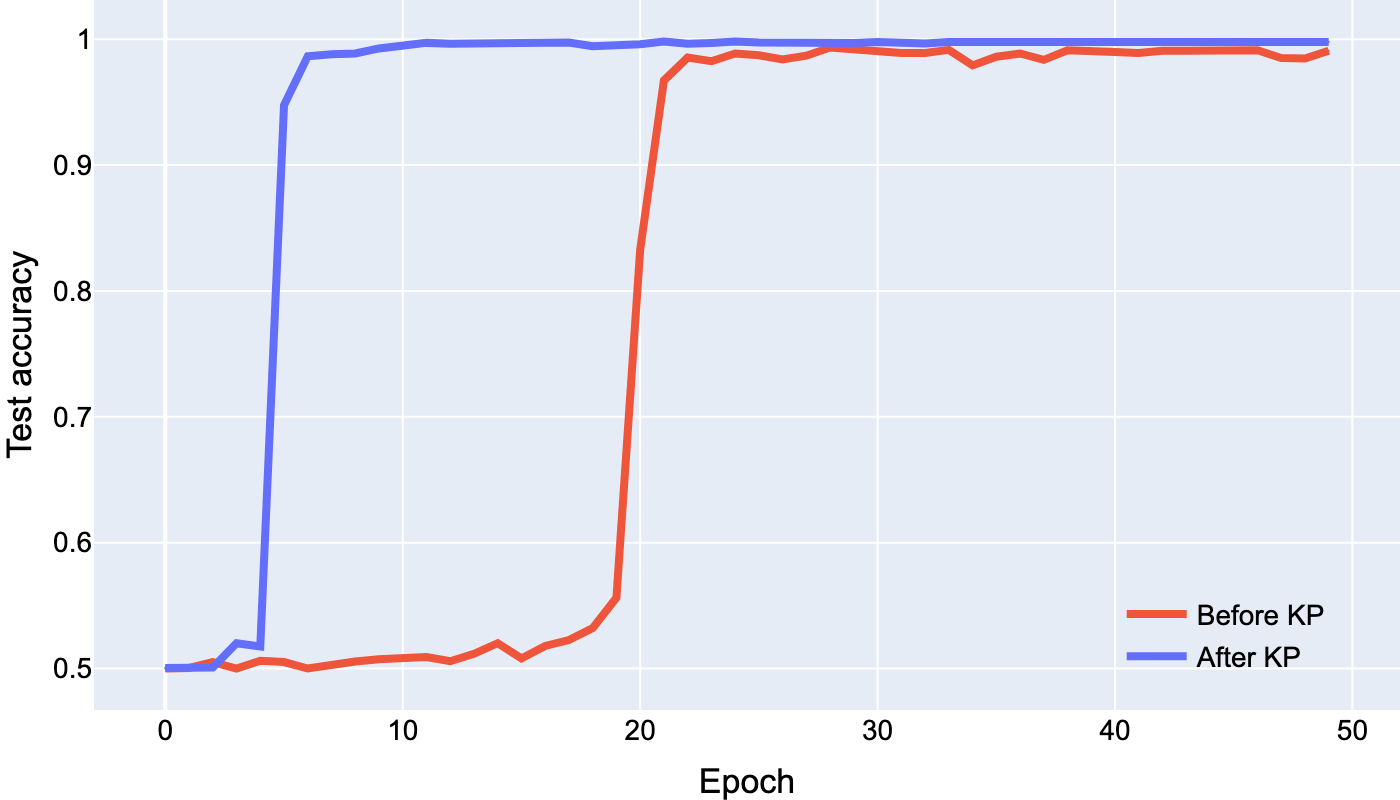}
	\caption{Training curve before and after applying kernel padding for the micro-object classification task. After kernel padding, ResNets capture the small perturbation difference more easily and faster.}
	\label{fig:traj}
\end{figure}

Experimental details such as training method and data augmentation are almost the same as in Section \ref{sec:Size_Test_on_the_Effective_Receptive_Field}. Here, to better see the intrinsic architectural differences, we did not use pre-trained weights. The number of epochs was set to 50. The observed training curve is shown in Figure \ref{fig:traj}. Initially, the test accuracy was around $50\%$, and the difference in perturbation was not captured. After a certain epoch, the test accuracy increased rapidly, and the difference in the micro-objects was captured with an accuracy of more than $99\%$. Here, the existing model without kernel padding required more epochs to capture the perturbation difference. In contrast, the model to which kernel padding is applied captured the perturbation difference faster.

To verify this more strictly, we measured the number of epoch where the test accuracy first exceeded $90\%$. If it did not exceed $90\%$ within 50 epochs, it was evaluated as 50 epochs. For ResNet-101 and its variant, five experiments were performed, and the average of the measured number of epochs was summarized (Table \ref{tab:epoch}). Even in the same training environment, after kernel padding, the difference in the micro-objects was captured 13-18 epochs faster. Thus, for some special tasks, pixel sensitivity imbalance is harmful to training.

\begin{table}[t!]
	\centering
	\begin{tabular}{l|r|r|r}
		\toprule
		\textbf{Model}    & \textbf{Before KP} & \textbf{After KP} & \textbf{Diff} \\
		\midrule
        ResNet-101        & 28.2               & \textbf{10.2}              & 18.0 $\downarrow$        \\
        Wide-ResNet-101-2 & 28.0               & \textbf{14.2}              & 13.8 $\downarrow$        \\
        ResNeXt-101-32x8d & 21.8               & \textbf{8.8}               & 13.0 $\downarrow$        \\
		\bottomrule
	\end{tabular}
	\caption{From five experiments, we computed the average number of epochs where the test accuracy exceeds $90\%$ for the micro-object classification task. After kernel padding, CNNs capture the small perturbation more easily and faster.}
	\label{tab:epoch}
\end{table}

\section{Conclusion}
\label{sec:Conclusion}

In this study, we investigated the behaviors of CNNs using effective receptive fields. First, we investigated the size of the receptive field. Contrary to popular belief, we found that the classification accuracy is not proportional to the size of the receptive field. In addition, we observed that the size of the effective receptive field saturates to a certain level even if the CNN becomes deeper. These observations suggest that we need to reconsider when controlling the size of the receptive field. Second, the pixels contributing to the output were investigated through the effective receptive field of the output. We discovered that in modern ResNets, the contribution to the output is different for each pixel. It was identified that the cause of this pixel sensitivity imbalance lies in the use of an odd-sized kernel with stride 2. To solve this, kernel padding was proposed. We quantitatively evaluated pixel sensitivity imbalance through two indices and found that pixel sensitivity imbalance decreases after kernel padding. We discussed that although the pixel sensitivity imbalance is a helpful feature for general vision tasks, it is a harmful bug for some tasks. These behaviors of CNNs should be understood and considered by practitioners.


\bibliography{aaai22}

\begin{thebibliography}{49}
\providecommand{\natexlab}[1]{#1}

\bibitem[{Araujo, Norris, and Sim(2019)}]{araujo2019computing}
Araujo, A.; Norris, W.; and Sim, J. 2019.
\newblock Computing receptive fields of convolutional neural networks.
\newblock \emph{Distill}, 4(11): e21.

\bibitem[{Azulay and Weiss(2019)}]{azulay2018deep}
Azulay, A.; and Weiss, Y. 2019.
\newblock Why do deep convolutional networks generalize so poorly to small
  image transformations?
\newblock \emph{Journal of Machine Learning Research}, 20: 1--25.

\bibitem[{Chattopadhay et~al.(2018)Chattopadhay, Sarkar, Howlader, and
  Balasubramanian}]{chattopadhay2018grad}
Chattopadhay, A.; Sarkar, A.; Howlader, P.; and Balasubramanian, V.~N. 2018.
\newblock Grad-cam++: Generalized gradient-based visual explanations for deep
  convolutional networks.
\newblock In \emph{2018 IEEE winter conference on applications of computer
  vision (WACV)}, 839--847. IEEE.

\bibitem[{Cohen and Welling(2016)}]{cohen2016group}
Cohen, T.; and Welling, M. 2016.
\newblock Group equivariant convolutional networks.
\newblock In \emph{International conference on machine learning}, 2990--2999.
  PMLR.

\bibitem[{Fei-Fei, Fergus, and Perona(2004)}]{fei2004learning}
Fei-Fei, L.; Fergus, R.; and Perona, P. 2004.
\newblock Learning generative visual models from few training examples: An
  incremental bayesian approach tested on 101 object categories.
\newblock In \emph{2004 conference on computer vision and pattern recognition
  workshop}, 178--178. IEEE.

\bibitem[{Fu et~al.(2018)Fu, Gong, Wang, Batmanghelich, and Tao}]{fu2018deep}
Fu, H.; Gong, M.; Wang, C.; Batmanghelich, K.; and Tao, D. 2018.
\newblock Deep ordinal regression network for monocular depth estimation.
\newblock In \emph{Proceedings of the IEEE conference on computer vision and
  pattern recognition}, 2002--2011.

\bibitem[{Ghiasi, Lin, and Le(2018)}]{ghiasi2018dropblock}
Ghiasi, G.; Lin, T.-Y.; and Le, Q.~V. 2018.
\newblock DropBlock: A regularization method for convolutional networks.
\newblock \emph{Advances in Neural Information Processing Systems}, 31:
  10727--10737.

\bibitem[{Glorot, Bordes, and Bengio(2011)}]{glorot2011deep}
Glorot, X.; Bordes, A.; and Bengio, Y. 2011.
\newblock Deep sparse rectifier neural networks.
\newblock In \emph{Proceedings of the fourteenth international conference on
  artificial intelligence and statistics}, 315--323. JMLR Workshop and
  Conference Proceedings.

\bibitem[{He et~al.(2015)He, Zhang, Ren, and Sun}]{he2015delving}
He, K.; Zhang, X.; Ren, S.; and Sun, J. 2015.
\newblock Delving deep into rectifiers: Surpassing human-level performance on
  imagenet classification.
\newblock In \emph{Proceedings of the IEEE international conference on computer
  vision}, 1026--1034.

\bibitem[{He et~al.(2016)He, Zhang, Ren, and Sun}]{he2016deep}
He, K.; Zhang, X.; Ren, S.; and Sun, J. 2016.
\newblock Deep residual learning for image recognition.
\newblock In \emph{Proceedings of the IEEE conference on computer vision and
  pattern recognition}, 770--778.

\bibitem[{Huang et~al.(2017)Huang, Liu, Van Der~Maaten, and
  Weinberger}]{huang2017densely}
Huang, G.; Liu, Z.; Van Der~Maaten, L.; and Weinberger, K.~Q. 2017.
\newblock Densely connected convolutional networks.
\newblock In \emph{Proceedings of the IEEE conference on computer vision and
  pattern recognition}, 4700--4708.

\bibitem[{Iandola et~al.(2016)Iandola, Han, Moskewicz, Ashraf, Dally, and
  Keutzer}]{iandola2016squeezenet}
Iandola, F.~N.; Han, S.; Moskewicz, M.~W.; Ashraf, K.; Dally, W.~J.; and
  Keutzer, K. 2016.
\newblock SqueezeNet: AlexNet-level accuracy with 50x fewer parameters and< 0.5
  MB model size.
\newblock \emph{arXiv preprint arXiv:1602.07360}.

\bibitem[{Johnson, Alahi, and Fei-Fei(2016)}]{johnson2016perceptual}
Johnson, J.; Alahi, A.; and Fei-Fei, L. 2016.
\newblock Perceptual losses for real-time style transfer and super-resolution.
\newblock In \emph{European conference on computer vision}, 694--711. Springer.

\bibitem[{Kim, Lee, and Lee(2016)}]{kim2016accurate}
Kim, J.; Lee, J.~K.; and Lee, K.~M. 2016.
\newblock Accurate image super-resolution using very deep convolutional
  networks.
\newblock In \emph{Proceedings of the IEEE conference on computer vision and
  pattern recognition}, 1646--1654.

\bibitem[{Krizhevsky, Sutskever, and Hinton(2012)}]{krizhevsky2012imagenet}
Krizhevsky, A.; Sutskever, I.; and Hinton, G.~E. 2012.
\newblock Imagenet classification with deep convolutional neural networks.
\newblock \emph{Advances in neural information processing systems}, 25:
  1097--1105.

\bibitem[{Long, Shelhamer, and Darrell(2015)}]{long2015fully}
Long, J.; Shelhamer, E.; and Darrell, T. 2015.
\newblock Fully convolutional networks for semantic segmentation.
\newblock In \emph{Proceedings of the IEEE conference on computer vision and
  pattern recognition}, 3431--3440.

\bibitem[{Loshchilov and Hutter(2016)}]{loshchilov2016sgdr}
Loshchilov, I.; and Hutter, F. 2016.
\newblock Sgdr: Stochastic gradient descent with warm restarts.
\newblock \emph{arXiv preprint arXiv:1608.03983}.

\bibitem[{Luo et~al.(2016)Luo, Li, Urtasun, and Zemel}]{luo2016understanding}
Luo, W.; Li, Y.; Urtasun, R.; and Zemel, R. 2016.
\newblock Understanding the effective receptive field in deep convolutional
  neural networks.
\newblock In \emph{Proceedings of the 30th International Conference on Neural
  Information Processing Systems}, 4905--4913.

\bibitem[{Ma et~al.(2018)Ma, Zhang, Zheng, and Sun}]{ma2018shufflenet}
Ma, N.; Zhang, X.; Zheng, H.-T.; and Sun, J. 2018.
\newblock Shufflenet v2: Practical guidelines for efficient cnn architecture
  design.
\newblock In \emph{Proceedings of the European conference on computer vision
  (ECCV)}, 116--131.

\bibitem[{Newville et~al.(2016)Newville, Stensitzki, Allen, Rawlik, Ingargiola,
  and Nelson}]{newville2016lmfit}
Newville, M.; Stensitzki, T.; Allen, D.~B.; Rawlik, M.; Ingargiola, A.; and
  Nelson, A. 2016.
\newblock LMFIT: Non-linear least-square minimization and curve-fitting for
  Python.
\newblock \emph{Astrophysics Source Code Library}, ascl--1606.

\bibitem[{Odena, Dumoulin, and Olah(2016)}]{odena2016deconvolution}
Odena, A.; Dumoulin, V.; and Olah, C. 2016.
\newblock Deconvolution and checkerboard artifacts.
\newblock \emph{Distill}, 1(10): e3.

\bibitem[{Paszke et~al.(2017)Paszke, Gross, Chintala, Chanan, Yang, DeVito,
  Lin, Desmaison, Antiga, and Lerer}]{paszke2017automatic}
Paszke, A.; Gross, S.; Chintala, S.; Chanan, G.; Yang, E.; DeVito, Z.; Lin, Z.;
  Desmaison, A.; Antiga, L.; and Lerer, A. 2017.
\newblock Automatic differentiation in pytorch.

\bibitem[{Paszke et~al.(2019)Paszke, Gross, Massa, Lerer, Bradbury, Chanan,
  Killeen, Lin, Gimelshein, Antiga et~al.}]{paszke2019pytorch}
Paszke, A.; Gross, S.; Massa, F.; Lerer, A.; Bradbury, J.; Chanan, G.; Killeen,
  T.; Lin, Z.; Gimelshein, N.; Antiga, L.; et~al. 2019.
\newblock Pytorch: An imperative style, high-performance deep learning library.
\newblock \emph{Advances in neural information processing systems}, 32:
  8026--8037.

\bibitem[{Pl{\"o}tz and Roth(2018)}]{plotz2018neural}
Pl{\"o}tz, T.; and Roth, S. 2018.
\newblock Neural Nearest Neighbors Networks.
\newblock \emph{Advances in Neural Information Processing Systems}, 31:
  1087--1098.

\bibitem[{Redmon et~al.(2016)Redmon, Divvala, Girshick, and
  Farhadi}]{redmon2016you}
Redmon, J.; Divvala, S.; Girshick, R.; and Farhadi, A. 2016.
\newblock You only look once: Unified, real-time object detection.
\newblock In \emph{Proceedings of the IEEE conference on computer vision and
  pattern recognition}, 779--788.

\bibitem[{Ren et~al.(2015)Ren, He, Girshick, and Sun}]{ren2015faster}
Ren, S.; He, K.; Girshick, R.; and Sun, J. 2015.
\newblock Faster r-cnn: Towards real-time object detection with region proposal
  networks.
\newblock \emph{Advances in neural information processing systems}, 28: 91--99.

\bibitem[{Ronneberger, Fischer, and Brox(2015)}]{ronneberger2015u}
Ronneberger, O.; Fischer, P.; and Brox, T. 2015.
\newblock U-net: Convolutional networks for biomedical image segmentation.
\newblock In \emph{International Conference on Medical image computing and
  computer-assisted intervention}, 234--241. Springer.

\bibitem[{Russakovsky et~al.(2015)Russakovsky, Deng, Su, Krause, Satheesh, Ma,
  Huang, Karpathy, Khosla, Bernstein et~al.}]{russakovsky2015imagenet}
Russakovsky, O.; Deng, J.; Su, H.; Krause, J.; Satheesh, S.; Ma, S.; Huang, Z.;
  Karpathy, A.; Khosla, A.; Bernstein, M.; et~al. 2015.
\newblock Imagenet large scale visual recognition challenge.
\newblock \emph{International journal of computer vision}, 115(3): 211--252.

\bibitem[{Sandler et~al.(2018)Sandler, Howard, Zhu, Zhmoginov, and
  Chen}]{sandler2018mobilenetv2}
Sandler, M.; Howard, A.; Zhu, M.; Zhmoginov, A.; and Chen, L.-C. 2018.
\newblock Mobilenetv2: Inverted residuals and linear bottlenecks.
\newblock In \emph{Proceedings of the IEEE conference on computer vision and
  pattern recognition}, 4510--4520.

\bibitem[{Selvaraju et~al.(2017)Selvaraju, Cogswell, Das, Vedantam, Parikh, and
  Batra}]{selvaraju2017grad}
Selvaraju, R.~R.; Cogswell, M.; Das, A.; Vedantam, R.; Parikh, D.; and Batra,
  D. 2017.
\newblock Grad-cam: Visual explanations from deep networks via gradient-based
  localization.
\newblock In \emph{Proceedings of the IEEE international conference on computer
  vision}, 618--626.

\bibitem[{Shi et~al.(2020)Shi, Guo, Jiang, Wang, Shi, Wang, and Li}]{shi2020pv}
Shi, S.; Guo, C.; Jiang, L.; Wang, Z.; Shi, J.; Wang, X.; and Li, H. 2020.
\newblock Pv-rcnn: Point-voxel feature set abstraction for 3d object detection.
\newblock In \emph{Proceedings of the IEEE/CVF Conference on Computer Vision
  and Pattern Recognition}, 10529--10538.

\bibitem[{Shrikumar, Greenside, and Kundaje(2017)}]{shrikumar2017learning}
Shrikumar, A.; Greenside, P.; and Kundaje, A. 2017.
\newblock Learning important features through propagating activation
  differences.
\newblock In \emph{International Conference on Machine Learning}, 3145--3153.
  PMLR.

\bibitem[{Silberman and Guadarrama(2016)}]{silberman2016tensorflow}
Silberman, N.; and Guadarrama, S. 2016.
\newblock TensorFlow-Slim image classification model library.
\newblock \emph{URL https://github.
  com/tensorflow/models/tree/master/research/slim}.

\bibitem[{Simonyan, Vedaldi, and Zisserman(2013)}]{simonyan2013deep}
Simonyan, K.; Vedaldi, A.; and Zisserman, A. 2013.
\newblock Deep inside convolutional networks: Visualising image classification
  models and saliency maps.
\newblock \emph{arXiv preprint arXiv:1312.6034}.

\bibitem[{Singh and Davis(2018)}]{singh2018analysis}
Singh, B.; and Davis, L.~S. 2018.
\newblock An analysis of scale invariance in object detection snip.
\newblock In \emph{Proceedings of the IEEE conference on computer vision and
  pattern recognition}, 3578--3587.

\bibitem[{Smilkov et~al.(2017)Smilkov, Thorat, Kim, Vi{\'e}gas, and
  Wattenberg}]{smilkov2017smoothgrad}
Smilkov, D.; Thorat, N.; Kim, B.; Vi{\'e}gas, F.; and Wattenberg, M. 2017.
\newblock Smoothgrad: removing noise by adding noise.
\newblock \emph{arXiv preprint arXiv:1706.03825}.

\bibitem[{Springenberg et~al.(2015)Springenberg, Dosovitskiy, Brox, and
  Riedmiller}]{springenberg2014striving}
Springenberg, J.; Dosovitskiy, A.; Brox, T.; and Riedmiller, M. 2015.
\newblock Striving for Simplicity: The All Convolutional Net.
\newblock In \emph{ICLR (workshop track)}.

\bibitem[{Srinivas and Fleuret(2018)}]{srinivas2018knowledge}
Srinivas, S.; and Fleuret, F. 2018.
\newblock Knowledge transfer with jacobian matching.
\newblock In \emph{International Conference on Machine Learning}, 4723--4731.
  PMLR.

\bibitem[{Srivastava et~al.(2014)Srivastava, Hinton, Krizhevsky, Sutskever, and
  Salakhutdinov}]{srivastava2014dropout}
Srivastava, N.; Hinton, G.; Krizhevsky, A.; Sutskever, I.; and Salakhutdinov,
  R. 2014.
\newblock Dropout: a simple way to prevent neural networks from overfitting.
\newblock \emph{The journal of machine learning research}, 15(1): 1929--1958.

\bibitem[{Su, Vargas, and Sakurai(2019)}]{su2019one}
Su, J.; Vargas, D.~V.; and Sakurai, K. 2019.
\newblock One pixel attack for fooling deep neural networks.
\newblock \emph{IEEE Transactions on Evolutionary Computation}, 23(5):
  828--841.

\bibitem[{Sundararajan, Taly, and Yan(2017)}]{sundararajan2017axiomatic}
Sundararajan, M.; Taly, A.; and Yan, Q. 2017.
\newblock Axiomatic attribution for deep networks.
\newblock In \emph{International Conference on Machine Learning}, 3319--3328.
  PMLR.

\bibitem[{Sutskever et~al.(2013)Sutskever, Martens, Dahl, and
  Hinton}]{sutskever2013importance}
Sutskever, I.; Martens, J.; Dahl, G.; and Hinton, G. 2013.
\newblock On the importance of initialization and momentum in deep learning.
\newblock In \emph{International conference on machine learning}, 1139--1147.
  PMLR.

\bibitem[{Szegedy et~al.(2015)Szegedy, Liu, Jia, Sermanet, Reed, Anguelov,
  Erhan, Vanhoucke, and Rabinovich}]{szegedy2015going}
Szegedy, C.; Liu, W.; Jia, Y.; Sermanet, P.; Reed, S.; Anguelov, D.; Erhan, D.;
  Vanhoucke, V.; and Rabinovich, A. 2015.
\newblock Going deeper with convolutions.
\newblock In \emph{Proceedings of the IEEE conference on computer vision and
  pattern recognition}, 1--9.

\bibitem[{Tan and Le(2019)}]{tan2019efficientnet}
Tan, M.; and Le, Q. 2019.
\newblock Efficientnet: Rethinking model scaling for convolutional neural
  networks.
\newblock In \emph{International Conference on Machine Learning}, 6105--6114.
  PMLR.

\bibitem[{Tsai et~al.(2018)Tsai, Hung, Schulter, Sohn, Yang, and
  Chandraker}]{tsai2018learning}
Tsai, Y.-H.; Hung, W.-C.; Schulter, S.; Sohn, K.; Yang, M.-H.; and Chandraker,
  M. 2018.
\newblock Learning to adapt structured output space for semantic segmentation.
\newblock In \emph{Proceedings of the IEEE conference on computer vision and
  pattern recognition}, 7472--7481.

\bibitem[{Wah et~al.(2011)Wah, Branson, Welinder, Perona, and
  Belongie}]{wah2011caltech}
Wah, C.; Branson, S.; Welinder, P.; Perona, P.; and Belongie, S. 2011.
\newblock The caltech-ucsd birds-200-2011 dataset.

\bibitem[{Xie et~al.(2017)Xie, Girshick, Doll{\'a}r, Tu, and
  He}]{xie2017aggregated}
Xie, S.; Girshick, R.; Doll{\'a}r, P.; Tu, Z.; and He, K. 2017.
\newblock Aggregated residual transformations for deep neural networks.
\newblock In \emph{Proceedings of the IEEE conference on computer vision and
  pattern recognition}, 1492--1500.

\bibitem[{Zagoruyko and Komodakis(2016)}]{zagoruyko2016wide}
Zagoruyko, S.; and Komodakis, N. 2016.
\newblock Wide Residual Networks.
\newblock In \emph{British Machine Vision Conference 2016}. British Machine
  Vision Association.

\bibitem[{Zhang(2019)}]{zhang2019making}
Zhang, R. 2019.
\newblock Making convolutional networks shift-invariant again.
\newblock In \emph{International conference on machine learning}, 7324--7334.
  PMLR.

\end{thebibliography}

\end{document}